\documentclass[lettersize,journal]{IEEEtran}
\usepackage{amsmath,amsfonts}
\usepackage{algorithmic}
\usepackage{algorithm}
\usepackage{array}
\usepackage[caption=false,font=normalsize,labelfont=sf,textfont=sf]{subfig}
\usepackage{textcomp}
\usepackage{stfloats}
\usepackage{url}
\usepackage{verbatim}
\usepackage{graphicx}
\usepackage{cite}
\usepackage{tabularx}
\usepackage{multirow}
\usepackage{tabularx}
\usepackage{booktabs}

\begin{document}

\title{TRACE-Bot: Detecting Emerging LLM-Driven Social Bots via Implicit Semantic Representations and AIGC-Enhanced Behavioral Patterns}

\author{
    Zhongbo Wang,
    Zhiyu Lin,
    Zhu Wang,
    Haizhou Wang\textsuperscript{*}%
    \thanks{This work was supported in part by the National Natural Science Foundation of China (NSFC) under Grant 62572331.}%
    \thanks{Zhongbo Wang is  with the School of Cyber Science and Engineering, Sichuan University, Sichuan 610207, China, and also with the School of Software and Microelectronics, Peking University, Beijing 102600, China (e-mail: zhongbowang25@stu.pku.edu.cn).}%
    \thanks{Zhiyu Lin is with the College of Computer Science, Sichuan University, Sichuan 610065, China (e-mail: zhiyulin@stu.scu.edu.cn).}%
    \thanks{Zhu Wang is with the Law School, Sichuan University, Sichuan 610207, China (e-mail: wangzhu@scu.edu.cn).}%
    \thanks{Haizhou Wang is with the School of Cyber Science and Engineering, Sichuan University, Sichuan 610207, China (e-mail: whzh.nc@scu.edu.cn).
    }%
    \thanks{* Corresponding author: whzh.nc@scu.edu.cn.}%
}

\IEEEpubid{}

\maketitle

\begin{abstract}
Large Language Model-driven (LLM-driven) social bots pose a growing threat to online discourse by generating human-like content that evades conventional detection. Existing methods suffer from limited detection accuracy due to over-reliance on single-modality signals, insufficient sensitivity to the specific generative patterns of Artificial Intelligence-Generated Content (AIGC), and a failure to adequately model the interplay between linguistic patterns and behavioral dynamics. To address these limitations, we propose TRACE-Bot, a unified dual-channel framework that jointly models implicit semantic representations and AIGC-enhanced behavioral patterns. TRACE-Bot constructs fine-grained representations from heterogeneous sources, including personal information data, interaction behavior data and tweet data. A dual-channel architecture captures linguistic representations via a pretrained language model and behavioral irregularities via multidimensional activity features augmented with signals from state-of-the-art (SOTA) AIGC detectors. The fused representations are then classified through a lightweight prediction head. Experiments on two public LLM-driven social bot datasets demonstrate SOTA performance, achieving accuracies of 98.46\% and 97.50\%, respectively. The results further indicate strong robustness against advanced bot strategies, highlighting the effectiveness of jointly leveraging implicit semantic representations and AIGC-enhanced behavioral patterns for emerging LLM-driven social bot detection.
\end{abstract}

\begin{IEEEkeywords}
Social Bots, Large Language Models, Implicit Semantic Representations, Behavioral Sequences, Artificial Intelligence-Generated Content
\end{IEEEkeywords}

\section{Introduction}
\subsection{Background}
\IEEEPARstart{W}{ith} the deep integration of Artificial Intelligence (AI) and social media, public opinion dissemination is shifting from human-driven to human-machine coexistence. Central to this transformation are social bots: automated accounts designed to mimic humans and influence discourse on salient topics\cite{ref1}. Empirical studies indicate that 9\%-15\% of active Twitter accounts exhibit bot-like characteristics\cite{ref2}, significantly impacting information dynamics.

Recent advances in NLP \cite{ref3} and Large Language Models (LLMs) \cite{ref4,ref5} have catalyzed LLM-driven social bots \cite{ref6,ref7} that dynamically adapt linguistic and behavioral patterns to convincingly emulate humans \cite{ref8}. This sophistication invalidates the static features, behavioral stationarity, and linguistic diversity assumptions of conventional detection methods \cite{ref9,ref10,ref11}, causing notable performance degradation in SOTA models \cite{ref12, ref13}. Consequently, robust paradigms explicitly modeling the adaptive and temporally evolving nature of these bots are urgently needed.

\subsection{Related Work}
Existing social bot detection methods can be divided into two categories based on the underlying generation mechanisms of the bots: traditional social bot detection methods and LLM-driven social bot detection methods.

Traditional social bot detection has evolved from early rule-based or heuristic methods relying on behavioral or account indicators \cite{ref14,ref15,ref16, ref17} to Machine Learning (ML) techniques \cite{ref1,ref18,ref19,ref20,ref21}. More recently, Deep Learning (DL) and neural network-based methods \cite{ref22,ref23,ref24,ref25,ref26} have shifted focus toward automatic representation learning from raw textual, behavioral, and graph data, significantly reducing the need for manual feature engineering. Building on this, LLMs-based methods \cite{ref27,ref28,ref29} now leverage advanced contextual reasoning to enhance performance in adversarial settings.

Conversely, LLM-driven social bots generate highly human-like content with adaptive interaction strategies, posing severe challenges to current detectors. Most existing methods rely heavily on linguistic patterns or isolated content features, lacking effective fusion of heterogeneous multimodal data and failing to jointly model textual semantics and behavioral dynamics. Consequently, these approaches struggle to distinguish LLM-driven social bots from authentic users, leading to degraded accuracy. Addressing these limitations necessitates novel paradigms specifically tailored for LLM-driven social bots to ensure the security and trustworthiness of digital ecosystems.

\subsection{Challenges}
In recent years, research on social bots has primarily focused on traditional social bots, while studies on detecting the new generation LLM-driven social bots are relatively scarce and face several challenges as follows:

First, they exhibit limited capability in modeling and effectively fusing heterogeneous multimodal information. Most methods rely on unimodal inputs (e.g., text or behavior alone) or adopt simplistic fusion strategies, failing to capture the complex interplay between non-human semantics and LLM-generated behavioral patterns. Second, current models primarily focus on surface-level textual or statistical behavioral features and lack sensitivity to intrinsic AIGC characteristics, such as generation regularities, distributional biases, and latent semantic traces introduced by LLMs. Third, semantic representations and behavioral patterns are often modeled independently, ignoring their intrinsic correlations. This decoupled paradigm hinders the alignment of implicit semantic representations with behavioral signatures, thereby limiting detection performance in LLM-driven social bot scenarios.

\subsection{Contributions}
To address the aforementioned challenges in detecting LLM-driven social bots, this study proposes a novel detection framework named TRACE-Bot, which integrates implicit semantic representations and AIGC-enhanced behavioral patterns within a unified dual-channel architecture. The main contributions of this study are summarized as follows:

\begin{itemize}
    \item To overcome limitations in fusing heterogeneous multimodal data, we propose a novel feature processing approach tailored for LLM-driven bots. By integrating personal information data, interaction behavior data, and tweet data, our method improves feature-level detection accuracy and precision by approximately 0.4\%.

    \item To address the insensitivity of existing features to AIGC patterns, we construct AIGC-enhanced behavioral patterns by integrating SOTA detectors, enabling efficient identification of LLM-driven social bots. To the best of our knowledge, this is the first work leveraging AIGC detection for social bot detection.

    \item We propose a behavior sequence construction method that integrates multidimensional interaction features into time-structured sequences. These sequences characterize bot patterns in operational modes, activity rhythms, and interaction strategies, enabling precise differentiation between automated accounts and human users.

    \item To overcome the limitation of decoupled text-behavior modeling, we propose a dual-channel collaborative framework. It jointly integrates features: a GPT-2-based text channel captures linguistic style and coherence, while an MLP-based behavior channel mines anomalous patterns and LLMs traces from normalized features. Experiments show a peak accuracy of 98.46\% in detecting LLM-driven social bots. To the best of our knowledge, this is the first work exploring text-behavior dual-channel fusion for social bot detection.
\end{itemize}

\subsection{Organization}
The remainder of this paper is organized as follows: Section II reviews related work; Section III details the TRACE-Bot architecture; Section IV describes the experimental design, implementation, and results analysis; Section V discusses limitations; Section VI outlines future directions; and Section VII concludes with key findings and contributions.

\section{RELATED WORK}
\subsection{Rule-based or Simple Heuristic Methods}
Early studies employed rule-based or simple heuristic methods exploiting observable patterns via four key features. \textit{Account metadata} (e.g., creation time, follower ratios) reveals abnormal bot lifespans\cite{ref14}. \textit{Activity levels} detect high-frequency interactions, achieving up to 70\% social bot detection accuracy\cite{ref15}. \textit{Content characteristics} identify repetitive, template-like structures with limited vocabulary\cite{ref16}. \textit{Network-level features} capture dense clustering and low reciprocity\cite{ref17}. 

Rule-based or heuristic methods rely on predefined thresholds and manually designed rules, offering advantages in simplicity and computational efficiency. However, as social bots increasingly mimic human behaviors, the differences between bots and human users become subtle and intentionally concealed. Consequently, approaches based on static features and fixed rules are easily evaded and exhibit limited generalization ability, resulting in reduced detection performance against advanced social bots.

\subsection{Machine Learning-based Methods}
Compared with rule-based or heuristic methods, ML-based methods improve detection accuracy and flexibility by learning discriminative patterns from multiple feature sources, including linguistic signals, behavioral dynamics, and network structures\cite{ref18}. Supervised classifiers (e.g., DT, SVM, RF) trained on labeled datasets effectively distinguish social bots from humans\cite{ref19,ref20}, while semi-supervised graph representation learning leverages user relationships and label propagation for high-precision detection\cite{ref21}. Furthermore, linguistically informed features (e.g., sentiment, lexical diversity) enhance adaptability to evolving strategies\cite{ref1}. 

ML-based methods typically rely on supervised classifiers trained on manually engineered features. While effective, their performance is highly dependent on feature design, which can be labor-intensive and may limit generalization to newly emerging LLM-driven social bot behaviors.

\subsection{Deep Learning and Neural Network-based Methods}
Deep Learning and Neural Network-based methods automate complex representation learning from raw data, reducing manual feature engineering. By jointly exploiting profiles, content, and metadata, these methods achieve superior accuracy and robustness over traditional methods. Yang et al.\cite{ref22} proposed \textit{FedACK}, a GAN-based framework with federated knowledge distillation for collaborative decentralized learning. Li et al.\cite{ref23} modeled homophilic and heterophilic connections via a tailored message-passing mechanism to capture structural heterogeneity. Qiao et al.\cite{ref24} presented \textit{BECE}, which mitigates the interference of camouflaged interactions by evaluating edge confidence via parameterized Gaussian distributions to filter unreliable connections in graph structures. Liu et al.\cite{ref25} introduced a community-aware spatiotemporal hypergraph contrastive learning framework to enhance node representations under semi-supervised settings. Lin et al.\cite{ref26} developed \textit{BotBR}, integrating behavioral extraction with attention-based feature fusion and an edge detector to optimize graph structural utilization. 

Deep Learning and Neural Network-based methods enable end-to-end learning of complex social bot-related patterns and significantly enhance detection performance. However, they still face challenges such as limited cross-modal representation capability and reliance on large-scale labeled data, which restrict their effectiveness when confronting highly human-like or rapidly evolving social bots.

\subsection{Large Language Models-based Methods}
Recent advances in LLMs have opened new opportunities for social bot detection due to their strong capabilities in semantic understanding and contextual reasoning. By leveraging pretrained language representations, these approaches aim to overcome limitations of earlier methods, such as constrained feature representation and dependence on extensive labeled data.

Zhou et al.\cite{ref27} applied Supervised Fine-Tuning (SFT) to adapt LLMs for social bot detection, enabling effective encoding of user textual sequences and improving performance when combined with downstream classifiers such as multilayer perceptrons. Feng et al.\cite{ref28} proposed an LLMs-based detector built on a heterogeneous mixture-of-experts framework, which adopts a divide-and-conquer strategy to process heterogeneous user information. Sundararajan et al.\cite{ref29} explored interpretable AI techniques, including LIME, SHAP, and Integrated Gradients, to analyze feature attribution in deep models and enhance the interpretability of detection decisions.

LLMs-based methods exploit the strong language modeling capability of LLMs to capture subtle patterns in AI-generated text, thereby improving the detection of highly human-like and sophisticated social bots.

\subsection{LLM-driven Social Bot Detection Methods}
Existing social bot detection methods have achieved considerable success against traditional social bots. Recently, several studies have begun to explore the detection of LLM-driven social bots. Duan et al.\cite{ref9} proposed \textit{LLM-BotGuard}, which integrates pattern-guided feature extraction, a mixture-of-experts module, and a graph module to capture both distinctive characteristics of LLM-driven social bots and common patterns shared with traditional social bots. Building upon this work, Duan et al.\cite{ref10} further introduced \textit{BotDMM}, which extracts discriminative features from user content and structural attributes and adopts a joint training strategy with self-supervised learning to enhance the differentiation among human users, traditional social bots, and LLM-driven social bots.

Existing research on detecting social bots generated by LLMs remains limited in quantity. Moreover, most existing studies rely on unimodal or bimodal signals, without attempting to construct fine-grained features from multimodal data sources. They also tend to model textual semantics and user behavioral patterns in isolation. In addition, the role of AIGC detection in identifying LLM-driven social bots has largely been overlooked.

\section{Methodology}

\begin{figure*}[!t]
\centering
\includegraphics[width=\textwidth]{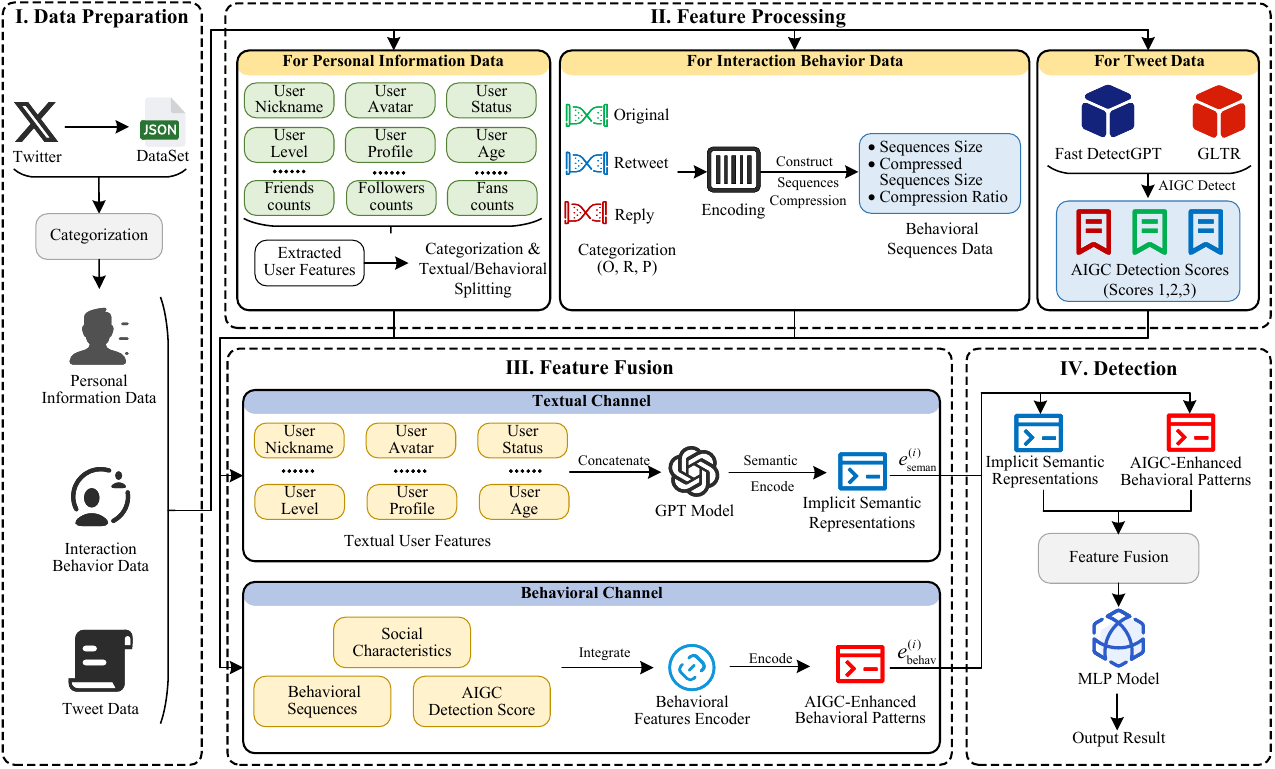} 
\caption{The proposed social bot detection model in this paper.}
\label{fig:1} 
\end{figure*}

To address challenges posed by LLM-driven social bots, we propose TRACE-Bot\footnote{\url{https://github.com/forthgoer1/TRACE-Bot}}, a dual-channel model integrating implicit semantic representations with AIGC-enhanced behavioral patterns (Fig. \ref{fig:1}). The architecture comprises four core modules: \textit{Data Preparation}, \textit{Feature Processing}, \textit{Feature Fusion}, and \textit{Detection}.

The \textit{Data Preparation} module handles data collection and preprocessing to ensure structured inputs. The \textit{Feature Processing} module extracts multidimensional textual and behavioral features, capturing semantic traces and interaction patterns. Subsequently, the \textit{Feature Fusion} module integrates these heterogeneous representations to enhance discriminative expressiveness. Finally, the \textit{Detection} module utilizes the fused vectors for precise identification and classification of LLM-driven social bots.

\subsection{Data Preparation Module}
This study relies exclusively on publicly available datasets. While numerous open-source Twitter social bot datasets exist (e.g., Twibot-22\cite{ref32}, Twibot-20\cite{ref33}, Gilani-2017\cite{ref34}, Cresci-Stock-2018\cite{ref35}), most target traditional social bots, rendering them unsuitable for detecting LLM-driven social bots. Consequently, we utilize Fox8-23\footnote{\url{https://zenodo.org/records/8035290}} and BotSim-24\footnote{\url{https://github.com/QQQQQQBY/BotSim/tree/main/BotSim-24-Dataset}}\cite{ref36}, the relatively limited datasets specifically addressing LLM-driven social bots with the scale and multimodal richness required by our model. Fox8-23 comprises over 9,000 accounts and 12,000 tweets of real-world LLM-generated content, balanced with verified human accounts from Botometer-Feedback-2019\footnote{\url{https://botometer.osome.iu.edu/bot-repository/datasets.html}}, Gilani-17\cite{ref34}, Midterm-2018\cite{ref37}, and Varol-Icwsm\cite{ref2}. BotSim-24\cite{ref36} offers a large-scale benchmark based on the \textit{BotSim} framework, recording multi-round interactions and evolving social relationships in controlled environments. For training and evaluation, we apply a unified preprocessing procedure to organize raw records into structured user-level representations, extracting features from three complementary modalities: personal information data, interaction behavior data, and tweet data. These multimodal signals serve as inputs for the proposed framework's \textit{Feature Processing Module}, \textit{Feature Fusion Module} and \textit{Detection Module}.

\subsection{Feature Processing Module}
The \textit{Feature Processing Module} performs modality-specific feature engineering on the three data modalities produced by the \textit{Data Preparation Module}. Its goal is to transform raw multimodal inputs into structured representations that capture discriminative signals of LLM-driven social bots.

\subsubsection{Personal Information Data}
For personal information data, we structure account-level attributes that reflect observable profile configurations and platform metadata, enabling differentiation between authentic users and LLM-driven social bots. Specifically, we construct a user profile feature vector covering six categories:

\begin{itemize}
    \item \textit{User Identity Information}: Basic account attributes displayed on the Twitter profile page, including username, display name, screen name, account creation time, and whether team management features are enabled.

    \item \textit{Profile Configuration}: Profile customization attributes, such as the self-written bio, embedded website URLs, declared location, and the presence or type of profile and banner images.

    \item \textit{Engagement Metrics}: Quantitative indicators of platform activity, including follower count, followee count, list membership count, total tweets, total likes, and relationship status (e.g., mutual following or blocking).

    \item \textit{Verification and Permissions}: Account verification status and platform-granted capabilities, such as verification badges and special permissions (e.g., translation or professional features).

    \item \textit{Privacy and Security Settings}: Privacy-related configurations, including whether the account is protected and whether geolocation tagging is disabled.

    \item \textit{Language and Timezone Preferences}: User-configured locale settings, including preferred interface language and declared timezone.
\end{itemize}

These profile-derived features capture both static identity attributes and configuration patterns, providing informative signals for identifying non-human behaviors associated with LLM-driven social bots.

\subsubsection{Interaction Behavior Data}
To capture distinctive behavioral patterns on social platforms, the \textit{Feature Processing Module} constructs behavior sequences tailored for social bot detection. Compared with human users, LLM-driven social bots often exhibit more regular and coordinated interaction patterns, such as high-frequency posting, retweeting, or replying within short time intervals. This temporal regularity makes behavioral sequences a useful signal for identifying automated accounts.

Behavior sequence construction follows three steps.

First, each tweet is categorized into one of three interaction types:

\begin{itemize}
    \item \textit{Original}: a tweet authored by the user.
    \item \textit{Retweet}: a repost of another user's content.
    \item \textit{Reply}: a response to another user's post.
\end{itemize}

This classification uses three metadata fields: \texttt{in\_reply\_to\_status\_id}, \texttt{in\_reply\_to\_user\_id}, and \texttt{retweeted\_status}. A tweet is labeled as \textit{Reply} if either \texttt{in\_reply\_to\_status\_id} or \texttt{in\_reply\_to\_user\_id} is non-null; otherwise, it is labeled as \textit{Retweet} if \texttt{retweeted\_status} is non-null; if neither condition holds, it is classified as \textit{Original}.

Second, each tweet type is mapped to a sequence symbol according to Equation (\ref{eqn:1}), where $O$, $R$, and $P$ denote Original, Retweet, and Reply, respectively.

\begin{equation}
\label{eqn:1}
B_{\text{type}}^{3}(x) = 
\begin{cases}
O, & \text{if } x = \text{Original}, \\
R, & \text{if } x = \text{Retweet}, \\
P, & \text{if } x = \text{Reply}.
\end{cases}
\end{equation}

Third, the symbols of all tweets are concatenated in chronological order to form a behavior sequence for each user.

To quantify behavioral regularity, we compress each sequence using the \texttt{zlib} library\footnote{\url{http://www.zlib.net/}}. Two auxiliary features are then extracted: the compressed sequence length and the compression ratio (compressed size divided by original size). Repetitive and structured sequences, which are often produced by automated accounts, tend to achieve higher compression ratios than the more diverse sequences of human users, making these metrics effective indicators for detecting LLM-driven social bots.

\subsubsection{Tweet Data}
For tweet data, the \textit{Feature Processing Module} leverages two SOTA AIGC detection models, Fast DetectGPT\cite{ref30} and GLTR\cite{ref31}, to assess the likelihood that a user’s posts were produced by LLMs. This provides a crucial discriminative signal for identifying LLM-driven social bots. Meanwhile, we acknowledge that AIGC detectors are imperfect and vulnerable to adversarial prompting, potentially yielding false classifications. To mitigate this, we employ two SOTA models for cross-validation to minimize errors. Furthermore, given that modern LLMs outputs can still evade detection, we treat these signals as auxiliary probabilistic features rather than definitive indicators.

Our method is grounded in the hypothesis that \textit{tweets posted by LLM-driven social bots are generated by AI systems}, and thus can be distinguished from human-authored content through AIGC detection. Specifically, we observe that while LLMs can mimic stylistic and emotional cues learned from large-scale corpora, their outputs may still reveal subtle inconsistencies in coherence, contextual understanding, and originality. Such discrepancies make AIGC detection feasible and provide probabilistic signals for identifying LLM-driven social bots. After applying Fast DetectGPT and GLTR to all tweets published by a user, the module aggregates the resulting detection scores to construct user-level features. Specifically, for each detector we compute several summary statistics, including the mean, standard deviation, maximum, minimum, and the proportion of tweets whose generation probability exceeds a predefined quantile threshold. These aggregated metrics form a comprehensive AIGC profile for each user and serve as key input features in the multimodal detection framework.

\subsection{Feature Fusion Module}
The \textit{Feature Fusion Module} adopts a dual-channel architecture based on GPT-2 and a Multilayer Perceptron (MLP) to encode and fuse feature vectors from personal information, interaction behavior, and tweet data. As these inputs comprise both textual and numerical attributes, naive transformations such as hashing or one-hot encoding for text, or converting numerical features into text, may lead to the loss of critical discriminative information. Moreover, interaction behavior and tweet-derived features are inherently numerical (e.g., statistics, probability distributions, compression ratios), while only a subset of profile features is textual. To effectively preserve this heterogeneity, we design a dual-channel fusion framework that separates inputs into a \textit{text channel} and a \textit{behavior channel}, each processed with a dedicated encoding strategy.

\subsubsection{Textual Channel}
We leverage GPT-2 to model semantic patterns in profile text, moving beyond manual feature engineering to mine implicit semantic representations, which are latent contextual patterns learned by pretrained language models. Our selection of GPT-2 is driven by three strategic considerations: First, its open-source nature ensures complete reproducibility, a fundamental requirement in information security and digital forensics. Second, utilizing a mature, standardized backbone isolates our method's contributions to ensure fair comparison and reproducibility. Finally, our method is orthogonal to the backbone architecture, enabling seamless transfer to newer models without modification. Notably, despite this conservative choice, our method achieves SOTA performance.

First, multiple textual attributes ${{T}_{i}}=\{{{t}_{1}},{{t}_{2}},...,{{t}_{n}}\}$ from the user $i$ profile are cleaned and concatenated into a unified input sequence ${X_{text}^{(i)}}$, as defined in Equation (\ref{eqn:2}), where $L$ denotes the maximum sequence length; sequences are truncated or padded accordingly.

\begin{equation}
\label{eqn:2}
X_{\text{text}}^{(i)} = \mathrm{Concat}\bigl(t_{1}^{(i)}, t_{2}^{(i)}, \dots, t_{n}^{(i)}\bigr) \in \mathbb{R}^{L}
\end{equation}

This sequence ${X_{text}^{(i)}}$ is then fed into the GPT-2 encoder to obtain contextualized hidden representations ${{H}^{(i)}}$, as shown in Equation (\ref{eqn:3}).

\begin{equation}
\label{eqn:3}
H^{(i)} = \mathrm{GPT}\bigl(\mathrm{Tokenize}(X_{\text{text}}^{(i)})\bigr) \in \mathbb{R}^{L \times d_h}
\end{equation}

Since GPT-2 lacks an explicit $[CLS]$ token, we adopt mean pooling with attention mask to derive a global text embedding that aggregates semantics from all valid tokens while excluding padding positions. Specifically, the final implicit semantic representations ${e_{seman}^{(i)}}$ is computed as in Equation (\ref{eqn:4}), where ${{m}_{t}}\in \{0,1\}$ is the attention mask and $\varepsilon $ is a small constant for numerical stability. This mechanism adaptively weights informative tokens, yielding a robust representation of user linguistic style and potential AI-generation traces.

\begin{equation}
\label{eqn:4}
e_{\text{seman}}^{(i)} = 
\frac{\sum_{t=1}^{L} H^{(i)} m_t}
     {\sum_{t=1}^{L} m_t + \varepsilon}
\end{equation}

\subsubsection{Behavioral Channel}
We feed the social characteristics, generated behavioral sequences, and detection scores from AIGC models into a Multilayer Perceptron (MLP) for non-linear transformation, forming AIGC-enhanced behavioral patterns to uncover LLM-driven or non-human characteristics. 

Let ${X_{behav}^{(i)}\in {{\mathbb{R}}^{{{d}_{b}}}}}$ denote the concatenated vector of numerical profile attributes, interaction behavior features, and AIGC detection statistics for user $i$, where $d_b$ is the total feature dimension. We standardize ${\tilde{X}_{behav}^{(i)}}$ using \texttt{StandardScaler} based on the training set mean $\boldsymbol{\mu}$ and standard deviation $\boldsymbol{\sigma}$, as formalized in Equation (\ref{eqn:5}).

\begin{equation}
\label{eqn:5}
\tilde{X}_{\text{behav}}^{(i)} = \frac{X_{\text{behav}}^{(i)} - \mu}{\sigma}
\end{equation}

The standardized vector is then passed through an MLP with ReLU activation and Dropout regularization to enhance sensitivity to subtle behavioral anomalies, producing the AIGC-enhanced behavioral patterns ${e_{behav}^{(i)}}$, as defined in Equation (\ref{eqn:6}), where $\sigma (\cdot )=\operatorname{Re}LU(\cdot )$, ${{W}_{1}}\in {{\mathbb{R}}^{{{d}_{m}}\times {{d}_{b}}}}$ and $d_m$ is the output dimension of the MLP.

\begin{equation}
\label{eqn:6}
e_{\text{behav}}^{(i)} = \mathrm{Dropout}\bigl( \sigma( W_1 \tilde{X}_{\text{behav}}^{(i)} + b_1 ) \bigr) \in \mathbb{R}^{d_m}
\end{equation}

Finally, the two channel-specific embeddings are concatenated to form the joint multimodal representation ${{z}^{(i)}}$, as shown in Equation (\ref{eqn:7}), which serves as the input to the downstream detection module.

\begin{equation}
\label{eqn:7}
z^{(i)} = \bigl[ e_{\text{seman}}^{(i)} \,;\, e_{\text{behav}}^{(i)} \bigr] \in \mathbb{R}^{d_h + d_m}
\end{equation}

\subsection{Detection Module}
The \textit{Detection Module} constructs a discriminative classifier based on the joint representation ${{z}^{(i)}}$ from the \textit{Feature Fusion Module}. We employ a lightweight two-hidden-layer MLP to retain multimodal information while learning a decision boundary for binary classification. 

First, the input ${{z}^{(i)}}$ is projected into a 256-dimensional intermediate space via a fully connected layer, followed by ReLU activation to enhance expressiveness, as defined in Equation (\ref{eqn:8}). To mitigate overfitting, Dropout regularization with rate $p$ is applied immediately after, as shown in Equation (\ref{eqn:9}):

\begin{equation}
\label{eqn:8}
h_1 = \mathrm{ReLU}\bigl( W_1 z^{(i)} + b_1 \bigr), \quad W_1 \in \mathbb{R}^{256 \times d}
\end{equation}
\begin{equation}
\label{eqn:9}
\tilde{h}_1 = \mathrm{Dropout}(h_1; p)
\end{equation}

Subsequently, a second hidden layer compresses the representation to 128 dimensions, also followed by ReLU activation and Dropout, as formulated in Equations (\ref{eqn:10}) and (\ref{eqn:11}):

\begin{align}
h_2 &= \mathrm{ReLU}\bigl( W_2 \tilde{h}_1 + b_2 \bigr), \quad W_2 \in \mathbb{R}^{128 \times 256} \label{eqn:10} \\
\tilde{h}_2 &= \mathrm{Dropout}(h_2; p) \label{eqn:11}
\end{align}

Finally, the classification logits are generated through a linear output layer, as formulated in Equation (\ref{eqn:12}). These logits are then normalized into a posterior probability distribution via the Softmax function, as defined in Equation (\ref{eqn:13}), where $y=0$ denotes a human user and $y=1$ indicates an LLM-driven social bot:

\begin{equation}
\label{eqn:12}
o = W_3 \tilde{h}_2 + b_3 \in \mathbb{R}^{2}
\end{equation}
\begin{equation}
\label{eqn:13}
\hat{y} = \mathrm{Softmax}(o) = \bigl[ P(y=0 \mid z^{(i)}),\; P(y=1 \mid z^{(i)}) \bigr]
\end{equation}

Given the severe class imbalance in real-world social networks where bots constitute a minority, we adopt a class-weighted cross-entropy loss for optimization, as defined in Equation (\ref{eqn:14}). Here, the class weight $w_k$ is inversely proportional to the frequency of class $k$, a strategy that effectively enhances the model’s sensitivity and recall toward the minority class:

\begin{equation}
\label{eqn:14}
L_{cls} = - \sum_{k=0}^{1} w_k \cdot I(y = k) \cdot \log \hat{y}_k
\end{equation}

\section{Experiments}
In this section, we conduct extensive experiments and analyses on two datasets specifically designed for LLM-driven social bots. The results demonstrate that our proposed method effectively addresses the challenges associated with multimodal information integration and generalization.

\subsection{Experimental Settings}
\subsubsection{Experimental Environment}
All detection models in this study are implemented using PyTorch and Scikit-learn. Experiments are conducted on a dedicated server equipped with an Intel(R) Xeon(R) Platinum 8255C CPU, an NVIDIA T4 GPU with 32 GB of VRAM, and 32 GB of system memory. The software environment consists of Python v3.8.10, PyTorch v2.4.1, and CUDA v12.2.0.

\subsubsection{Datasets}
We evaluate our model on the public Fox8-23 and BotSim-24~\cite{ref36} datasets, collected in compliance with platform policies without private user data (statistics in Table~\ref{tab:1}). Both are split randomly into 60\% training, 20\% validation, and 20\% testing sets. To address class imbalance in BotSim-24, we apply random undersampling to the bot class in the training and validation subsets to match human counts, while preserving the test set's original distribution to reflect realistic deployment. This strategy mitigates training bias from majority-class dominance while ensuring evaluation reliability and comparability.

\begin{table}[!t]
\caption{Dataset Statistics}
\label{tab:1}
\centering
\footnotesize
\begin{tabularx}{\linewidth}{
>{\centering\arraybackslash}X
>{\centering\arraybackslash}X
>{\centering\arraybackslash}X
>{\centering\arraybackslash}X}
\toprule
\textbf{Dataset} & \textbf{Category} & \textbf{User Count} & \textbf{Tweet Count} \\
\midrule

\multirow{2}{*}{Fox8-23}
& Genuine Users & 1,140 & 218,245 \\
& Social Bots   & 1,140 & 149,783 \\
\cmidrule(lr){1-4}

\multirow{2}{*}{BotSim-24}
& Genuine Users & 1,000 & 129,722 \\
& Social Bots   & 1,906 & 129,728 \\
\bottomrule

\end{tabularx}
\end{table}

\subsubsection{Baseline Models}
We evaluate our method against 11 representative open-source social bot detection models, grouped into four categories: traditional Machine Learning, Deep Learning, Graph Neural Networks (GNNs) and Large Language Models-based approaches.

\begin{itemize}
    \item \textit{Traditional Machine Learning}: \textit{Pasricha et al.}\cite{ref38} introduces \textit{Digital DNA}, encoding behavioral patterns into symbolic sequences to capture temporal differences between bots and humans. \textit{BotRuler}\cite{ref39} is a rule-based detector leveraging heuristic features such as friend acquisition rate and posting frequency.

    \item \textit{Deep Learning}: \textit{Arin et al.}\cite{ref43} employs multiple LSTM modules to capture sequential behavior, while \textit{Mou et al.}\cite{ref39} combines handcrafted and deep learning features in a hybrid framework.

    \item \textit{Graph Neural Networks}: \textit{BotRGCN} variants (BotRGCN1, BotRGCN2, BotRGCN12)\cite{ref41} construct relational GCNs over user graphs with different modalities. \textit{BotMoE}\cite{ref42} uses a multimodal mixture-of-experts to fuse metadata, text, and network embeddings. \textit{UnDBot}\cite{ref44} and \textit{SEBot}\cite{ref45} leverage structural entropy and community-aware strategies for unsupervised detection. \textit{CACL}\cite{ref47} applies heterogeneous graph contrastive learning, and \textit{BotDGT}\cite{ref48} models dynamic temporal graphs.

    \item \textit{Large Language Models}: \textit{LMBot}\cite{ref46} injects graph knowledge into language models for bot detection. TRACE-Bot represents our proposed LLM-driven framework, achieving state-of-the-art performance.
\end{itemize}

\subsubsection{Evaluation Metrics}
We assess model performance using four standard metrics\cite{ref49}: Accuracy (ACC), Precision (PRE), Recall (REC), and F1-Score (F1). The model is trained for a maximum of 10 epochs with a batch size of 256, employing early stopping based on the validation F1-Score to mitigate overfitting.

\begin{table*}[!t]
\caption{Performance of Different Social Bot Detection Methods on Fox-8-23 and BotSim-24}
\label{tab:2}
\centering
\footnotesize
\resizebox{\textwidth}{!}{
\begin{tabular}{
>{\centering\arraybackslash}p{3.2cm}
>{\centering\arraybackslash}p{3.0cm}
*{8}{c}}

\toprule
\multirow{2}{*}{\textbf{Category}} & \multirow{2}{*}{\textbf{Model}} 
& \multicolumn{4}{c}{\textbf{Fox8-23}} 
& \multicolumn{4}{c}{\textbf{BotSim-24}} \\
\cmidrule(lr){3-6} \cmidrule(lr){7-10}
 &  & \textbf{ACC} & \textbf{PRE} & \textbf{REC} & \textbf{F1} 
 & \textbf{ACC} & \textbf{PRE} & \textbf{REC} & \textbf{F1} \\
\midrule

\multirow{2}{*}{\textbf{Traditional Machine Learning}} 
& Pasricha et al.\cite{ref38} 
& 0.8202 & 0.8687 & 0.7544 & 0.8075 
& 0.5300 & 0.5337 & 0.4500 & 0.4891 \\

& BotRuler\cite{ref40} 
& 0.5424 & 0.5223 & \textbf{0.9990} & 0.6860 
& 0.5000 & 0.5000 & \textbf{0.9990} & 0.6664 \\

\midrule
\multirow{2}{*}{\textbf{Deep Learning}} 
& Arin et al.\cite{ref43} 
& 0.9627 & 0.9451 & 0.9825 & 0.9634 
& 0.8150 & 0.8247 & 0.8000 & 0.8122 \\

& Mou et al.\cite{ref39} 
& 0.9620 & 0.9489 & 0.9766 & 0.9625 
& 0.6300 & 0.6186 & 0.8210 & 0.7050 \\

\midrule
\multirow{8}{*}{\textbf{Graph Neural Networks}}
& BotRGCN1\cite{ref41} 
& 0.6754 & 0.5894 & 0.9953 & 0.7404 
& 0.5675 & 0.8571 & 0.0957 & 0.1722 \\

& BotRGCN2\cite{ref41} 
& 0.9737 & 0.9587 & 0.9858 & 0.9721 
& 0.6550 & 0.6214 & 0.6809 & 0.6497 \\

& BotRGCN12\cite{ref41} 
& 0.8947 & 0.8178 & 0.9953 & 0.8979 
& 0.6915 & 0.6185 & \textbf{0.9990} & 0.7643 \\

& BotMoE\cite{ref42} 
& 0.9320 & 0.9218 & 0.9492 & 0.9353 
& 0.6625 & 0.7890 & 0.4343 & 0.5603 \\

& UnDBot\cite{ref44} 
& 0.5197 & 0.5101 & \underline{0.9982} & 0.6752 
& 0.5000 & 0.5000 & \textbf{0.9990} & 0.6664 \\

& SEBot\cite{ref45} 
& 0.9342 & 0.9360 & 0.9435 & 0.9398 
& 0.6000 & 0.8000 & 0.3636 & 0.5000 \\

& CACL\cite{ref47} 
& \underline{0.9803} & \underline{0.9784} & 0.9826 & \underline{0.9806} 
& \underline{0.9700} & \underline{0.9518} & \underline{0.9954} & \underline{0.9731} \\

& BotDGT\cite{ref48} 
& 0.5575 & 0.7175 & 0.5395 & 0.4303 
& 0.8417 & 0.8443 & 0.8400 & 0.8407 \\

\midrule
\multirow{2}{*}{\textbf{Large Language Models}}
& LMBot\cite{ref46} 
& 0.8969 & 0.9103 & 0.8826 & 0.8962 
& 0.4850 & 0.4326 & 0.4231 & 0.4278 \\

& TRACE-Bot
& \textbf{0.9846} & \textbf{0.9825} & 0.9868 & \textbf{0.9847} 
& \textbf{0.9750} & \textbf{0.9567} & 0.9950 & \textbf{0.9755} \\

\bottomrule
\end{tabular}}
\vspace{2pt}
\begin{flushleft}
\footnotesize \textit{Note:} The best performance is highlighted in \textbf{bold}, and the second-best performance is \underline{underlined}.
\end{flushleft}
\end{table*}

\subsection{Comparative Study}
Table \ref{tab:2} compares TRACE-Bot with baselines on Fox8-23 and BotSim-24. Key findings include:

\begin{itemize}
    \item TRACE-Bot consistently outperforms all baselines across most metrics. It achieves the best Accuracy, Precision, and F1-Score on Fox8-23. On BotSim-24, it surpasses the second-best method by +0.50 Accuracy, +0.49 Precision, and +0.24 F1-Score, demonstrating an optimal balance between detection accuracy and reliability.

    \item While methods like \textit{BotRuler}\cite{ref40}, \textit{UnDBot}\cite{ref44}, and \textit{BotRGCN12}\cite{ref41} achieve near-perfect Recall ($ = $0.9990), they suffer from critically low Precision ($ < $0.62), indicating excessive false positives due to over-labeling. In contrast, TRACE-Bot maintains high Recall while achieving 0.9825 Precision, offering a more practical solution for real-world deployment.

    \item Traditional social bots detectors (e.g., \textit{BotRGCN1}\cite{ref41}, \textit{SEBot}\cite{ref45}, \textit{LMBot}\cite{ref46}) and single-modality approaches (e.g., \textit{Pasricha et al.}\cite{ref38}, \textit{Arin et al.}\cite{ref43}) degrade significantly on these datasets, with some yielding F1-Scores below 0.60 or even 0.17 (e.g., \textit{BotRGCN1}\cite{ref41} on BotSim-24). This highlights the limitations of rigid behavioral or single-cue models against LLM-driven content. TRACE-Bot overcomes these limitations by jointly leveraging multimodal semantics, social topology, and dynamic behaviors, ensuring robust detection even in highly human-like scenarios.
\end{itemize}

\subsection{Ablation Study}
To systematically evaluate the effectiveness of TRACE-Bot, we conduct two sets of ablation studies: \textit{module ablation} and \textit{modality ablation}.

\begin{table}[!t]
\caption{Results of Module Ablation Study of Our Method}
\label{tab:3}
\centering
\footnotesize
\begin{tabularx}{\linewidth}{
>{\raggedright\arraybackslash}X
c c c c}
\toprule
\textbf{Module Settings} & \textbf{ACC} & \textbf{PRE} & \textbf{REC} & \textbf{F1} \\
\midrule

TRACE-Bot
& \textbf{0.9846} 
& \textbf{0.9825} 
& \textbf{0.9868} 
& \textbf{0.9847} \\

\quad - Textual Channel 
& 0.9561 & 0.9483 & 0.9649 & 0.9565 \\

\quad - Behavioral Channel
& 0.9189 & 0.8687 & 0.9868 & 0.9240 \\

\quad \parbox[t]{\dimexpr\linewidth-2em}{\hangindent=1em\hangafter=1 - Textual Channel and\\ Behavioral Channel}
& 0.9474 & 0.9359 & 0.9605 & 0.9481 \\

\bottomrule
\end{tabularx}
\vspace{2pt}
\begin{flushleft}
\footnotesize \textit{Note:} The best performance is highlighted in \textbf{bold}.
\end{flushleft}
\end{table}

\subsubsection{Module Ablation Study}
We assess three variants: removing the textual channel, the behavioral channel and both where the classifier operates on concatenated raw features.  All variants are trained and evaluated under identical dataset splits and hyperparameter settings. The results are summarized in Table \ref{tab:3}.

The full model yields optimal performance. Removing the textual channel reduces the F1-Score by 2.82 to 0.9565, underscoring the value of implicit semantic representations. Conversely, removing the behavioral channel causes a sharper 6.07 F1 drop to 0.9240, driven by a significant Precision decline to 0.8687 (Recall remains stable). This indicates that fine-grained behavioral features are critical for suppressing false positives among linguistically human-like bots. The variant without both channels performs marginally better than the behavior-only version but lags significantly behind the full model, confirming that neither modality alone suffices for capturing complex LLM-driven social bot characteristics.

Overall, the results demonstrate the necessity of TRACE-Bot’s dual-channel design: the GPT-2-based text encoder captures linguistic traces of social bots, while the MLP-based behavior encoder identifies behavioral anomalies associated with LLMs generation, and their integration improves the model’s discriminative capability.

\begin{table*}[!t]
\caption{Results of Modality Ablation Study of Our Method}
\label{tab:4}
\centering
\footnotesize
\setlength{\tabcolsep}{5pt}

\begin{tabular*}{\textwidth}{@{\extracolsep{\fill}} >{\raggedright\arraybackslash}p{7.5cm} c c c c c c c c}
\toprule
\multirow{2}{*}{\textbf{Modality Settings}} 
& \multicolumn{4}{c}{\textbf{Fox8-23}} 
& \multicolumn{4}{c}{\textbf{BotSim-24}} \\
\cmidrule(lr){2-5} \cmidrule(lr){6-9}
& \textbf{ACC} & \textbf{PRE} & \textbf{REC} & \textbf{F1} 
& \textbf{ACC} & \textbf{PRE} & \textbf{REC} & \textbf{F1} \\
\midrule

TRACE-Bot
& \textbf{0.9846} & \textbf{0.9825} & 0.9868 & \textbf{0.9847} 
& 0.9750 & 0.9567 & 0.9950 & 0.9755 \\

\quad - Personal Information Data 
& 0.9276 & 0.9372 & 0.9167 & 0.9268 
& 0.7825 & 0.7756 & 0.7950 & 0.7852 \\

\quad - Interaction Behavior Data
& 0.9737 & 0.9655 & 0.9825 & 0.9739 
& 0.9424 & 0.8968 & \textbf{0.9990} & 0.9455 \\

\quad - Tweet Data
& 0.9803 & 0.9740 & 0.9868 & 0.9804 
& 0.9845 & 0.9708 & \textbf{0.9990} & 0.9848 \\

\quad - Personal Information Data and Interaction Behavior Data
& 0.8224 & 0.7795 & 0.8991 & 0.8350 
& 0.7950 & 0.7950 & 0.7950 & 0.7950 \\

\quad - Personal Information Data and Tweet Data
& 0.8816 & 0.9350 & 0.8202 & 0.8738 
& 0.5500 & 0.5902 & 0.3600 & 0.4472 \\

\quad - Interaction Behavior Data and Tweet Data 
& \textbf{0.9846} & 0.9784 & \textbf{0.9912} & \textbf{0.9847} 
& \textbf{0.9900} & \textbf{0.9900} & 0.9900 & \textbf{0.9900} \\

\bottomrule
\end{tabular*}

\vspace{2pt}
\begin{flushleft}
\footnotesize \textit{Note:} The best performance is highlighted in \textbf{bold}.
\end{flushleft}

\end{table*}

\begin{figure*}[!t]
\centering
\includegraphics[width=\textwidth]{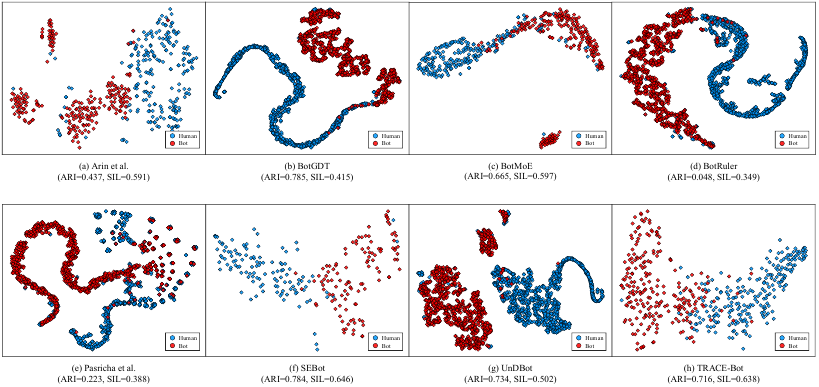} 
\caption{Results of representation learning study of our method on Fox8-23.}
\label{fig:2} 
\end{figure*}

\begin{figure*}[!t]
\centering
\includegraphics[width=\textwidth]{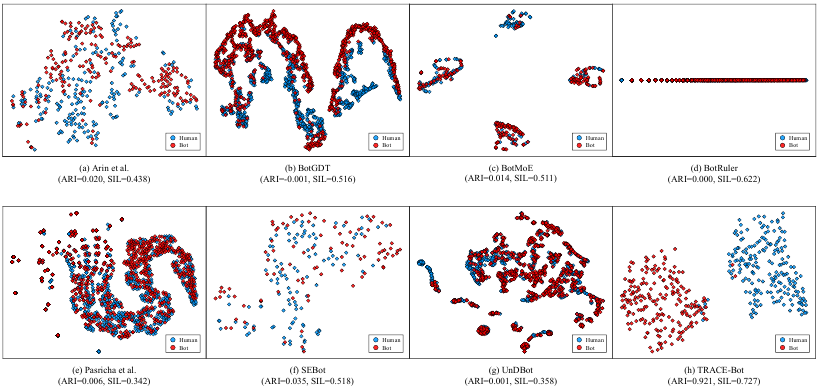} 
\caption{Results of representation learning study of our method on BotSim-24.}
\label{fig:3} 
\end{figure*}

\subsubsection{Modality Ablation Study}
We evaluate individual modality contributions by systematically removing inputs from the full labeled dataset (Table~\ref{tab:4}).

On Fox8-23, \textit{personal information data} alone matches the full model, indicating strong discriminative signals in metadata. However, on the challenging BotSim-24, unimodal approaches fail significantly: \textit{interaction behavior data} and \textit{tweet data} yield F1-Score of only 0.4472 and 0.7950, respectively. While personal information data remains robust, the full model is essential to maintain 0.9950 Recall and stabilize Precision against sophisticated camouflage.

Among bimodal combinations, \textit{personal information data} + \textit{interaction behavior data} achieves optimal balance (Fox8-23: 0.9804; BotSim-24: 0.9852), confirming that behavioral patterns effectively complement profile attributes. Conversely, \textit{personal information data} + \textit{tweet data} attains 0.9990 Recall on BotSim-24 but suffers lower Precision (0.8969) due to false positives without behavioral context. The \textit{interaction behavior data} + \textit{tweet data} pair performs worst (Fox8-23's F1-Score: 0.9268), underscoring the critical role of profile metadata.

In summary, TRACE-Bot’s robustness stems from the synergy of three components: profile metadata provides stable priors, interaction behavior captures anomalies, and GPT-2-encoded semantics identify LLM artifacts. Their integration ensures reliable detection across diverse scenarios.

\subsection{Representation Learning Study}
We evaluate the representation learning capability of TRACE-Bot via t-SNE visualization (Fig.\ref{fig:2} and Fig.\ref{fig:3}) and quantitative clustering metrics: Adjusted Rand Index (ARI) and Silhouette Index (SIL). All methods use identical hyperparameters for fair comparison.

Visually, TRACE-Bot demonstrates superior separability on both datasets. On Fox8-23, it forms distinct clusters for humans (blue) and bots (red). On the challenging BotSim-24, it yields two compact, nearly non-overlapping clusters, indicating effective modeling of behavioral and linguistic nuances.

Quantitative results corroborate these findings. On Fox8-23, TRACE-Bot achieves an ARI of 0.716 and SIL of 0.638, outperforming \textit{Arin et al.}\cite{ref43} (ARI=0.437) and \textit{BotMoE}\cite{ref42} (ARI=0.665). On BotSim-24, it shows even greater superiority with an ARI of 0.921 and SIL of 0.727, significantly surpassing the next best method, \textit{Mou et al.}\cite{ref39} (ARI=0.748). In contrast, baselines like \textit{BotRuler}\cite{ref40}, \textit{SEBot}\cite{ref45}, \textit{BotDGT}\cite{ref48}, and \textit{UnDBot}\cite{ref44} exhibit noticeable class overlap or limited intra-cluster compactness.

This advantage stems from TRACE-Bot's dual-channel fusion: the GPT-2 encoder captures semantic coherence and LLM artifacts, while the MLP encoder models anomalous behaviors. Together, they construct a highly discriminative representation space, ensuring robust inter-class separation and intra-cluster consistency for accurate social bot detection.

\begin{figure}[!t]
\centering
\includegraphics[width=3.5in]{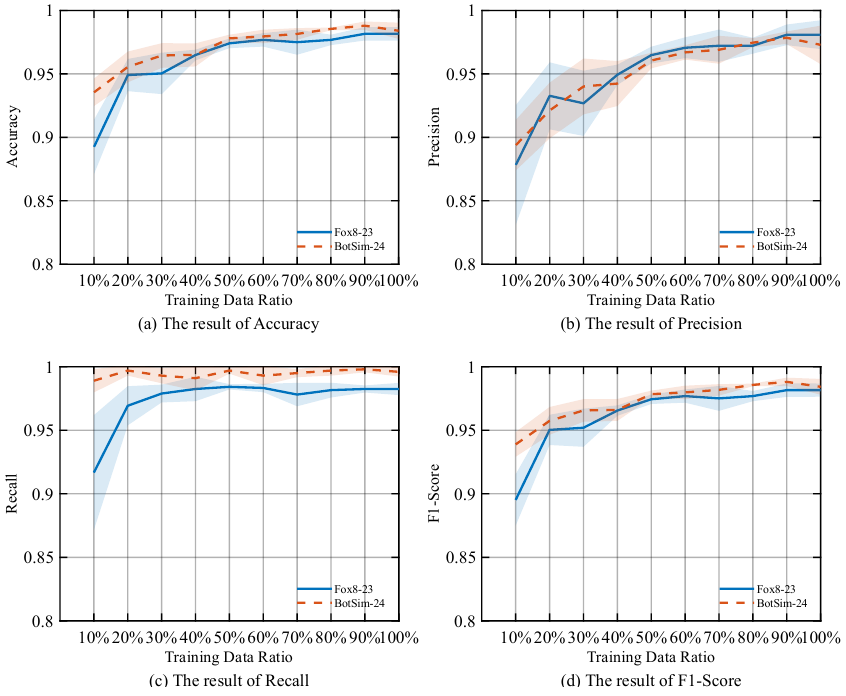}
\caption{Results of label efficiency study of our method.}
\label{fig:4}
\end{figure}

\subsection{Label Efficiency Study}

To systematically evaluate the data efficiency and generalization capability of TRACE-Bot under limited annotation budgets, a realistic constraint in large-scale social media applications, we conduct a controlled label efficiency study. In this experiment, the model architecture is fixed to the full three-modality version (Full Model) and trained using only a fraction (10\%–100\%) of the labeled training samples. For each sampling ratio, we perform 5-fold cross-validation with independent random splits to reduce sampling bias.

As shown in Fig.\ref{fig:4}, TRACE-Bot demonstrates strong data efficiency across different labeling ratios. On Fox8-23, the model achieves an F1-Score of 0.8952 using only 10\% of the labeled data, and performance nearly saturates at a 40\% labeling ratio. On the more challenging BotSim-24 dataset, TRACE-Bot reaches an F1-Score of 0.9656 with 30\% labels and approaches its peak performance (0.9857) at 80\% labeling. Notably, on BotSim-24, TRACE-Bot consistently maintains a high Recall ($\geq 0.9890 $) even under low labeling regimes, indicating strong sensitivity to bot instances with limited supervision. This robustness is attributed to the complementary multimodal representations, which help stabilize the decision boundary when labeled data are scarce.

\begin{figure*}[!t]
\centering
\includegraphics[width=\textwidth]{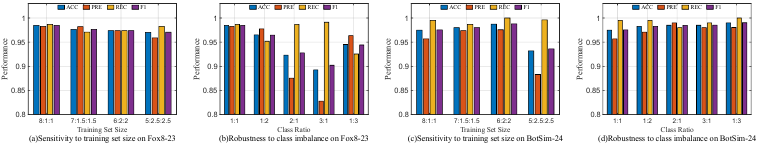} 
\caption{Results of robustness study of our method on Fox8-23 and BotSim-24.}
\label{fig:5} 
\end{figure*}

\begin{figure*}[!t]
\centering
\includegraphics[width=\textwidth]{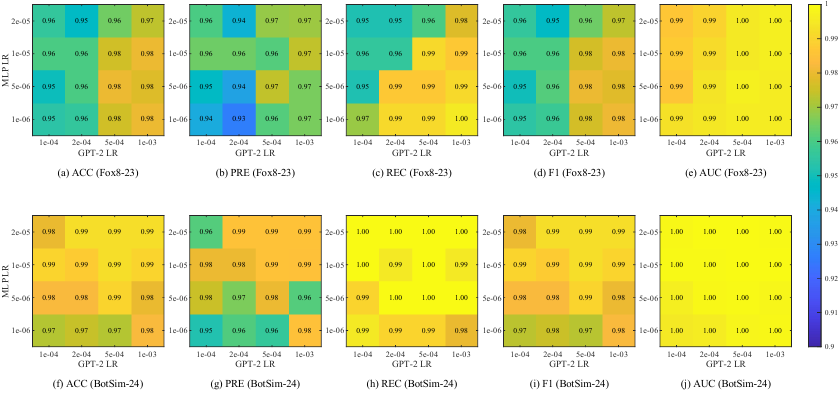} 
\caption{Results of hyperparameter study of our method on Fox8-23 and BotSim-24.}
\label{fig:6} 
\end{figure*}

\subsection{Robustness Study}
To evaluate the stability and generalization capability of TRACE-Bot under varying training conditions, we conduct two robustness experiments: sensitivity to training set size by varying the train/validation/test split ratios, and robustness to class imbalance by adjusting the bot-to-human ratio in the training set. Experiments are conducted on the Fox8-23 and BotSim-24.

\subsubsection{Sensitivity to Training Set Size}
We progressively reduce the training set proportion from 80\% to 50\%, correspondingly increasing the validation and test sets. As shown in Fig.\ref{fig:5}, TRACE-Bot maintains stable performance under reduced training data. On Fox8-23, when the training data is reduced to 50\%, the model still achieves 0.9702 Accuracy and 0.9705 F1-Score, with performance degradation below 1.5 percentage points. On BotSim-24, when the training ratio decreases to 50\%, the F1-Score drops from 0.9877 (under the 6:2:2 split) to 0.9361, while remaining above 0.93. Notably, under the standard 6:2:2 setting, TRACE-Bot achieves 0.9875 Accuracy, outperforming the configuration with an 80\% training ratio. These results indicate that TRACE-Bot maintains competitive performance even with substantially reduced training data.

\subsubsection{Robustness to Class Imbalance}
To simulate real-world label distribution shifts, we vary the bot-to-human ratio in the training set. As shown in Fig. \ref{fig:5}, on Fox8-23, when the ratio changes from 1:1 to 3:1, the F1-Score decreases from 0.9847 to 0.9022, while under the inverse imbalance (1:3) the F1-Score remains 0.9441. Although the 3:1 setting leads to a drop in Precision (0.8278), Recall consistently exceeds 99\%, indicating a conservative strategy that prioritizes high bot detection coverage. On BotSim-24, TRACE-Bot remains highly stable across all tested ratios (1:3 to 3:1), with the F1-Score consistently within [0.9800, 0.9900] and Accuracy fluctuating by less than 1.5\%. Under the 1:3 setting, the model achieves 0.9900 Accuracy and 1.0000 Recall. This suggests that TRACE-Bot can maintain strong discriminability even under significant label skew.

Overall, TRACE-Bot demonstrates strong robustness to both reduced training data and class imbalance, supporting its applicability in real-world social media environments.

\subsection{Hyperparameter Study}
To analyze TRACE-Bot's sensitivity to optimization settings, we evaluated performance across varying learning rates ($lr$) for the GPT-2 encoder and MLP encoder

As shown in Fig.\ref{fig:6}, on Fox8-23, optimal performance (0.9803 Accuracy, 0.9804 F1-Score, 0.9966 AUC) was achieved with MLP's $lr \in \{5\times10^{-6}, 1\times10^{-5}\}$ and GPT-2's $lr=1\times10^{-3}$; increasing MLP $lr$ to $2\times10^{-5}$ reduced the F1-Score to 0.9738, indicating instability with overly large rates. Conversely, on BotSim-24, TRACE-Bot demonstrated greater robustness: the optimal region spanned MLP's $lr \in [1\times10^{-5}, 2\times10^{-5}]$ and GPT-2's $lr \in [5\times10^{-4}, 1\times10^{-3}]$, achieving a peak F1-Score of 0.9926 and AUC of 0.9997. Notably, F1-Scores remained above 0.9850 across broader intervals (MLP: $[5\times10^{-6}, 2\times10^{-5}]$, GPT-2: $[2\times10^{-4}, 1\times10^{-3}]$). Synthesizing both datasets, we identified a consistent \textit{performance sweet spot} (MLP's $lr \in [5\times10^{-6}, 1\times10^{-5}]$, GPT-2's $lr \in [5\times10^{-4}, 1\times10^{-3}]$). Combining both datasets, we fixed MLP's $lr=1\times10^{-5}$ and GPT-2's $lr=5\times10^{-4}$ for subsequent experiments. 

Notably, across all tested learning rate intervals for both GPT-2 and MLP, TRACE-Bot demonstrates consistently strong performance on both datasets, with Accuracy exceeding 0.9500 in all cases. This confirms that the model possesses a remarkably large performance sweet spot, underscoring its robustness and superior design.

\begin{table}[!t]
\caption{Result of Comparison of Pretrained Language Models}
\label{tab:5}
\centering
\footnotesize
\setlength{\tabcolsep}{5pt}

\begin{tabular*}{\columnwidth}{@{\extracolsep{\fill}}ccccc}
\toprule
\textbf{Model Setting} & \textbf{ACC} & \textbf{PRE} & \textbf{REC} & \textbf{F1} \\
\midrule

BERT-base    
& 0.9737 & 0.9779 & 0.9693 & 0.9736 \\

RoBERTa-base 
& 0.9496 & 0.9362 & 0.9649 & 0.9503 \\

DistilGPT-2  
& 0.9320 & 0.9487 & 0.9781 & 0.9350 \\

DeBERTa-v3   
& 0.9605 & 0.9487 & 0.9737 & 0.9610 \\

GPT-2 (Proposed)
& \textbf{0.9846} & \textbf{0.9825} & \textbf{0.9868} & \textbf{0.9847} \\

\bottomrule
\end{tabular*}

\vspace{2pt}
\begin{flushleft}
\footnotesize \textit{Note:} The best performance is highlighted in \textbf{bold}.
\end{flushleft}

\end{table}

\subsection{Encoding Study}
To evaluate the textual semantic encoding module, we compare different Pretrained Language Models (PLMs) and input strategies for GPT-2.

\subsubsection{Comparison of Pretrained Language Models}
As shown in Table \ref{tab:5}, TRACE-Bot with \textit{GPT-2} achieves optimal performance on Fox8-23 (Accuracy: 0.9846, F1-Score: 0.9847), outperforming other PLMs. While \textit{BERT-base} and \textit{DeBERTa-v3} leverage bidirectional context, they yield lower F1-scores (0.9736 and 0.9610), likely due to reduced sensitivity to fine-grained anomalies (e.g., repetitiveness, logical discontinuities) indicative of LLM-generated content. Performance further degrades for \textit{RoBERTa-base} and \textit{DistilGPT-2}; notably, \textit{DistilGPT-2}'s low Precision (0.8956) suggests that parameter reduction compromises fine-grained semantic distinction. Conversely, \textit{GPT-2}'s autoregressive architecture and large-scale pretraining better capture latent linguistic irregularities in tweets and profiles.

\subsubsection{Comparison of Encoding Strategies}
We evaluate input strategies for GPT-2 by comparing direct raw-text concatenation against handcrafted prompt templates. As shown in Table \ref{tab:6}, all prompt-based variants degrade performance. The best prompt yields an F1-Score of 0.9673, roughly 1.7 below the prompt-free baseline, indicating that structured prompting is less effective than direct encoding for this task. This likely stems from social bot detection relying on subtle stylistic cues (e.g., lexical preferences, syntactic complexity, emotional tone); artificial prompts may introduce semantic noise or rigid formatting that dilutes these native signals. Since even minimal prompts underperform the raw-text baseline, our results highlight the superiority of direct raw-text encoding.

\begin{table*}[!t]
\caption{Result of Comparison of Prompt Encoding Strategies}
\label{tab:6}
\centering
\footnotesize
\setlength{\tabcolsep}{5pt}

\begin{tabular*}{\textwidth}{@{\extracolsep{\fill}}p{12cm}cccc}
\toprule
\textbf{Prompt Setting} & \textbf{ACC} & \textbf{PRE} & \textbf{REC} & \textbf{F1} \\
\midrule

``\{description\} \{location\} \{name\}'' 
& 0.9605 & 0.9526 & 0.9693 & 0.9609 \\

``This user's profile: name is \{name\}, located in \{location\}, and describes themselves as: \{description\}.'' 
& 0.9671 & 0.9610 & 0.9737 & 0.9673 \\

``Classify whether this social media account is a bot or human based on the following profile information: 
Name: \{name\}, Location: \{location\}, Profile: \{combined\_text\}'' 
& 0.9583 & 0.9524 & 0.9649 & 0.9586 \\

``Profile: \{combined\_text\}'' 
& 0.9605 & 0.9487 & 0.9737 & 0.9610 \\

\midrule
No Prompt (Proposed)
& \textbf{0.9846} & \textbf{0.9825} & \textbf{0.9868} & \textbf{0.9847} \\

\bottomrule
\end{tabular*}

\vspace{2pt}
\begin{flushleft}
\footnotesize \textit{Note:} The best performance is highlighted in \textbf{bold}.
\end{flushleft}

\end{table*}

\begin{figure}[!t]
\centering
\includegraphics[width=3.5in]{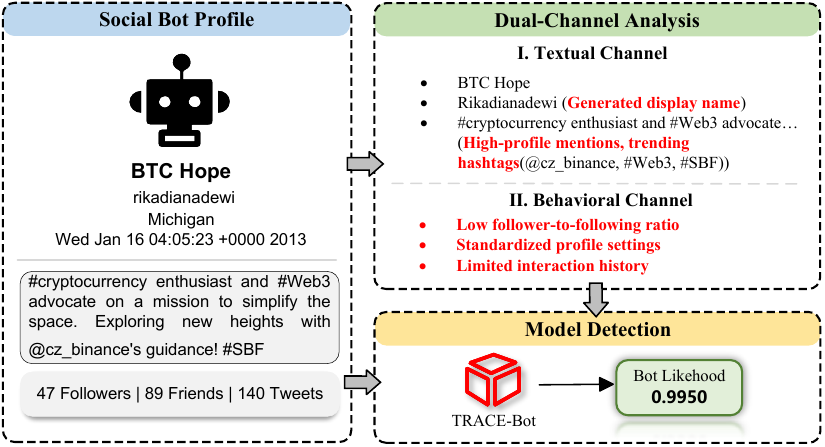}
\caption{The case study of our method.}
\label{fig:7}
\end{figure}

\subsection{Case Study}
To elucidate TRACE-Bot's dual-channel fusion, we analyze a representative Fox8-23 case correctly identified with high confidence (0.9950). Both single-modalities yielded consistent affirmative predictions: text-only (1.0000) and behavior-only (0.9898), indicating strong discriminative power in both features. As shown in Fig.~\ref{fig:7}, the user exhibits distinct LLM-driven signatures: a display name (BTC Hope) and random username (rikadianadewi) combining crypto keywords with arbitrary strings, alongside a bio aggregating trending hashtags (Web3, SBF) and mentions (@czbinance) via templated, impersonal phrasing. Concurrently, behavioral traits characterized by an extremely low follower-to-following ratio and sparse interactions align with typical bot patterns. The GPT-2 encoder captured subtle linguistic representations like unnatural fluency (over-smoothness), while the MLP behavior encoder effectively extracted signals from sparse data. This case underscores the dual-channel architecture's critical role, where synergistic multimodal leveraging ensures high accuracy and robustness through a dual-verification mechanism.

\section{Discussion}
Due to methodological scope and data constraints, this study has two limitations. First, TRACE-Bot's generalization across broader bot typologies requires further validation; the model is optimized for LLM-driven social bots, with performance on traditional rule-based or script-generated social bots not thoroughly evaluated. Second, the work focuses exclusively on English-language content on Twitter (now X), excluding other ecosystems (e.g., Facebook, Reddit, Weibo) and non-English contexts. Given significant cross-platform differences in user behavior, content, interaction, and moderation\cite{ref50, ref51}, TRACE-Bot may face adaptability challenges in cross-platform or multilingual settings.

\section{Future Work}
We outline three future directions. First, to extend model generalization, we will evaluate TRACE-Bot on traditional rule-based social bots by integrating diverse open-source datasets\cite{ref32,ref33,ref34,ref35} to enhance robustness across heterogeneous types. Second, we aim to build cross-platform and multilingual benchmarks by constructing an LLM-driven social bot dataset from Chinese platforms (e.g., Sina Weibo), which will allow us to localize and fine-tune TRACE-Bot for detecting emerging threats in Chinese social networks. Finally, we plan to incorporate multimodal generative content by extending our framework to handle images, videos, and audio through the integration of SOTA synthetic media detectors, thereby addressing next-generation bots that utilize multimodal deception.

\section{Conclusion}
This study proposes TRACE-Bot, a unified dual-channel framework for detecting LLM-driven social bots by jointly modeling implicit semantic representations and AIGC-enhanced behavioral patterns. By fusing high-dimensional multimodal representations via dual encoding channels where GPT-2 for implicit semantic representations and MLP for AIGC-enhanced behavioral patterns, TRACE-Bot effectively captures LLM-driven social bot characteristics. Experimental results demonstrate that TRACE-Bot significantly outperforms SOTA baselines, achieving accuracy scores of 98.46 and 97.50 on two datasets. Furthermore, ablation studies and hyperparameter analysis confirm the model's effectiveness and robustness against emerging LLM-driven social bot threats.

\section{Acknowledgements}
We gratefully acknowledge the guidance and support provided by our advisors for this work.

\vfill

\end{document}